\documentclass{article} 
\usepackage{times}
\usepackage{fullpage}
\usepackage{tikz}
\usepackage{algorithm}
\usepackage{algorithmic}
\usepackage{hyperref}
\usepackage{url}
\usepackage{natbib}
\usepackage{graphicx}
\usepackage{amsmath,amssymb,amsthm}

\DeclareMathOperator*{\argmin}{arg\,min}

\DeclareMathOperator*{\Peaks}{Peaks}
\DeclareMathOperator*{\Segments}{Segments}

\DeclareMathOperator*{\minimize}{minimize}
\newcommand{\sign}{\operatorname{sign}}
\newcommand{\RR}{\mathbb R}
\newcommand{\ZZ}{\mathbb Z}

\definecolor{noPeaks}{HTML}{F6F4BF}
\definecolor{peakStart}{HTML}{FFAFAF}
\definecolor{peakEnd}{HTML}{FF4C4C}
\definecolor{peaks}{HTML}{A445EE}
\newcommand{\JointHeuristic}{\textsc{JointZoom}}

\title{PeakSegJoint: fast supervised peak detection via joint
  segmentation of multiple count data samples}

\author{Toby Dylan Hocking toby.hocking@mail.mcgill.ca \\
Guillaume Bourque guil.bourque@mcgill.ca}

\begin{document}

\maketitle

\begin{abstract}
  Joint peak detection is a central problem when comparing samples in
  genomic data analysis, but current algorithms for this task are
  unsupervised and limited to at most 2 sample types. We propose
  PeakSegJoint, a new constrained maximum likelihood segmentation
  model for any number of sample types. To select the number of peaks
  in the segmentation, we propose a supervised penalty learning
  model. To infer the parameters of these two models, we propose to
  use a discrete optimization heuristic for the segmentation, and
  convex optimization for the penalty learning. In comparisons with
  state-of-the-art peak detection algorithms, PeakSegJoint achieves
  similar accuracy, faster speeds, and a more interpretable model
  with overlapping peaks that occur in exactly the same positions
  across all samples.
\end{abstract}

\tableofcontents

\newpage

\section{Introduction: Joint supervised peak detection in ChIP-seq data}


\begin{figure}[b!]
  \centering
  \includegraphics[width=\textwidth]{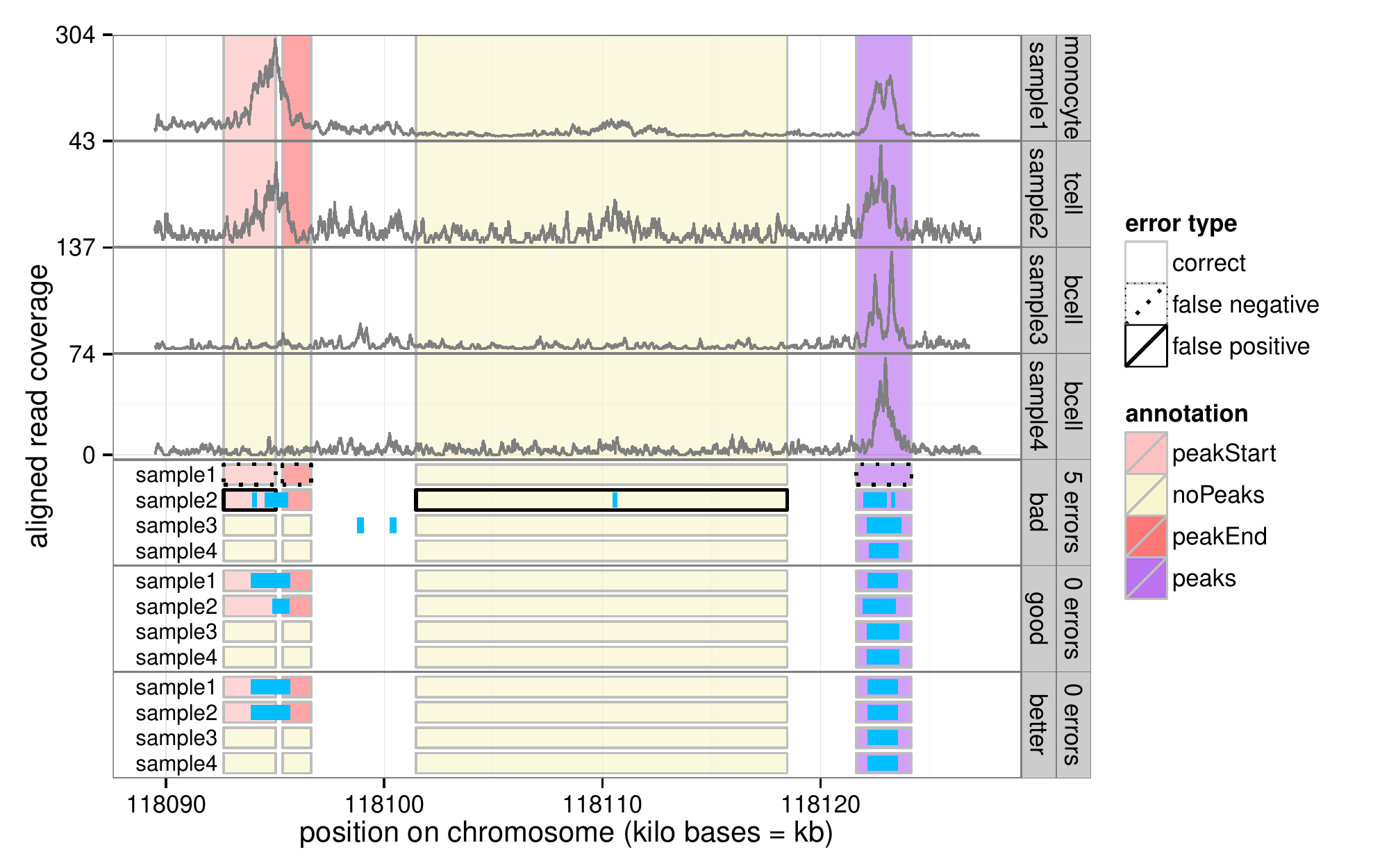}
  \vskip -0.5cm
  \caption{Labeled ChIP-seq coverage data for $S=4$ samples (top 4
    panels) with 3 peak models (bottom 3 panels).
    An ideal peak model minimizes the number of incorrect labels
    (false positives are too many peaks and false negatives are not
    enough peaks). The ``good'' model achieves 0 errors but the
    ``better'' model is more interpretable since overlapping peaks in
    different samples always occur in the exact same positions.}
  \label{fig:good-bad}
\end{figure}

Chromatin immunoprecipitation sequencing (ChIP-seq) is an experimental
technique used for genome-wide profiling of histone modifications and
transcription factor binding sites \citep{practical}. Each experiment
yields a set of sequence reads which are aligned to a reference
genome, and then the number of aligned reads are counted at each
genomic position. To compare samples at a given genomic position,
biologists visually examine coverage plots such as
Figure~\ref{fig:good-bad} for presence or absence of common ``peaks.''
In machine learning terms, a single sample over $B$ base positions is
a vector of non-negative count data $\mathbf z\in\ZZ_+^B$ and a peak
detector is a binary classifier $c:\ZZ_+^B\rightarrow\{0, 1\}^B$. The
positive class is peaks and the negative class is background
noise. Importantly, peaks and background occur in long contiguous
segments across the genome.

In this paper we use the supervised learning framework of
\citet{hocking2014visual}, who proposed to use labels obtained from
manual visual inspection of genome plots. For each labeled genomic
region $i\in\{1, \dots, n\}$ there is a set of count data $\mathbf
z_i$ and labels $L_i$ (``noPeaks,'' ``peaks,'' etc. as in
Figure~\ref{fig:good-bad}). These labels define a non-convex
annotation error function
\begin{equation}
  \label{eq:error}
  E[c(\mathbf z_i),  L_i] =
  \text{FP}[c(\mathbf z_i), L_i] +
  \text{FN}[c(\mathbf z_i), L_i]
\end{equation}
which counts the number of false positive (FP) and false negative (FN)
regions, so it takes values in the non-negative integers. In this
framework the goal of learning is to find a peak detection algorithm
$c$ that minimizes the number of incorrect labels (\ref{eq:error}) on
a test set of data:
\begin{equation}
  \label{eq:min_error}
  \minimize_c \sum_{i\in\text{test}} E[c(\mathbf z_i),  L_i].
\end{equation}

In practical situations we have $S>1$ samples, and a matrix $\mathbf
Z\in\ZZ_+^{B\times S}$ of count data. For example in
Figure~\ref{fig:good-bad} we have $S=4$ samples. In these data we are
not only interested in accurate peak detection in individual samples,
but also in detecting the differences between samples. In particular,
an ideal model of a multi-sample data set is a \textbf{joint} peak
detector with identical overlapping peak positions across samples (the
``better'' model in Figure~\ref{fig:good-bad}).

\subsection{Contributions and organization}

The main contribution of this paper is the \ref{PeakSegJoint} model
for supervised joint peak detection. Unlike previous methods
(Section~\ref{sec:related}), it is the first joint peak detector that
explicitly models any number of sample types
(Section~\ref{sec:models}). A secondary contribution is the
\JointHeuristic\ heuristic segmentation algorithm, and a supervised
learning algorithm for efficiently selecting the \ref{PeakSegJoint}
model complexity (Section~\ref{sec:algorithms}).  We show test error
and timing results in Section~\ref{sec:results}, provide a discussion
in Section~\ref{sec:discussion}, and propose some future research
directions in Section~\ref{sec:conclusions}.

\section{Related work}
\label{sec:related}

\subsection{Single-sample ChIP-seq peak detectors}

There are many different unsupervised peak detection algorithms
\citep{evaluation2010, rye2010manually, chip-seq-bench}, and the
current state-of-the-art is the supervised \ref{PeakSeg} model of
\citet{HOCKING-PeakSeg}. 

In the multi-sample setting of this paper, each of these algorithms
may be applied independently to each sample. The drawback of these
methods is that the peaks do not occur in the same positions across
each sample, so it is not straightforward to determine differences
between samples.

\subsection{Methods for several data sets}

This paper is concerned with the particular setting of analyzing
several samples (perhaps of different cell types) of the same
experiment type. For example, Figure~\ref{fig:good-bad} shows 2 bcell
samples, 1 monocyte sample, and 1 tcell sample (all of the H3K4me3
experiment type). As far as we know, there are no existing algorithms
that can explicitly model these data.

The most similar peak detectors in the bioinformatics literature model
several samples of the same cell type. For example, the JAMM algorithm
of \citet{JAMM} can analyze several samples, but is limited to a
single cell type since it assumes that each sample is a replicate with
the exact same peak pattern. So to analyze the data of
Figure~\ref{fig:good-bad} one would have to run JAMM 3 times (once for
each cell type), and the resulting peaks would not be the same across
cell types. Another example is the PePr algorithm of \citet{PePr},
which can model either one or two cell types, but is unsuitable for
analysis of three or more cell types. In contrast, \ref{PeakSegJoint}
is for data from several samples without limit on the number of cell
types.

Although not the subject of this paper, there are several algorithms
designed for the analysis of data from several different ChIP-seq
experiments \citep{chromhmm,segway,jmosaics}. In these models, the
input is one sample (e.g. monocyte cells) with different experiments
such as H3K4me3 and H3K36me3. Also,
\citet{hierarchical-joint} proposed a model for one sample with both
ChIP-seq and ChIP-chip data. In contrast, 
\ref{PeakSegJoint} takes as input several samples
(e.g. monocyte, tcell, bcell) for one experiment such as H3K4me3.

\newpage

\section{Models}\label{sec:models}

We begin by summarizing the single-sample \ref{PeakSeg} model, and
then introduce the multi-sample \ref{PeakSegJoint} model.

\subsection{PeakSeg: finding the most likely $0,\dots,p_{\text{max}}$
  peaks in a single sample}

This section describes the \ref{PeakSeg} model of
\citet{HOCKING-PeakSeg}, which is the current state-of-the-art peak
detection algorithm in the McGill benchmark data set of
\citet{hocking2014visual}. 

Given a single sample profile $\mathbf z\in\ZZ_+^B$ of aligned read counts
on $B$ bases, and a maximum number of peaks $p_{\text{max}}$, the
PeakSeg model for the mean vector $\mathbf m\in\RR^B$ with $p\in\{0,
\dots, p_{\text{max}}\}$ peaks is
\begin{align}
  \label{PeakSeg}
  \mathbf{\tilde m}^p(\mathbf z)  =
    \argmin_{\mathbf m\in\RR^{B}} &\ \ 
    \text{PoissonLoss}(\mathbf m, \mathbf z) 
    \tag{\textbf{PeakSeg}}
\\
    \text{such that} &\ \  \Peaks(\mathbf m)=p, \label{peaks=p} \\
     \forall j\in\{1, \dots, B\}, &\ \ P_j(\mathbf m) \in\{0, 1\},
    \label{eq:peak_constraint}
\end{align}
where the Poisson loss function is
\begin{equation}\label{eq:rho}
  \text{PoissonLoss}(\mathbf m, \mathbf z)= \sum_{j=1}^B m_j - z_j \log m_j.
\end{equation} 
Note that the optimization objective of minimizing the Poisson loss is
equivalent to maximizing the Poisson likelihood. The model complexity
(\ref{peaks=p}) is the number of peaks
\begin{equation}
  \Peaks(\mathbf m)=(\Segments(\mathbf m)-1)/2,
\end{equation}
which is a function of the number of piecewise constant segments
\begin{equation}
  \Segments(\mathbf m)=1+\sum_{j=2}^B I(m_j \neq m_{j-1}).
\end{equation}
Finally, the peak indicator at base $j$ is defined as the cumulative sum of
signs of changes
\begin{equation}
  \label{eq:peaks}
  P_j(\mathbf m) = \sum_{k=2}^j \sign( m_{k} - m_{k-1} ),
\end{equation}
with $P_1(\mathbf m)=0$ by convention. The constraint
(\ref{eq:peak_constraint}) means that the peak indicator $P_j(\mathbf
m)\in\{0, 1\}$ can be used to classify each base $j\in\{1,\dots B\}$
as either background noise $P_j(\mathbf m)=0$ or a peak $P_j(\mathbf
m)=1$.  In other words, $\mathbf{\tilde m}^p(\mathbf z)$ is a
piecewise constant vector that changes up, down, up, down (and not up,
up, down). Thus the even numbered segments are interpreted as peaks,
and the odd numbered segments are interpreted as background noise.

Note that since \ref{PeakSeg} is defined for a single sample, it may
be independently applied to each sample in data sets such as
Figure~\ref{fig:good-bad}. However, any overlapping peaks in different
samples will not necessarily occur in the same positions. In the next
section we fix this problem by proposing the more interpretable
multi-sample \ref{PeakSegJoint} model.

\newpage 

\subsection{PeakSegJoint: finding the most likely common peak in
  $0,\dots, S$ samples}

\begin{figure}[b!]
  \centering
  \includegraphics[width=\textwidth]{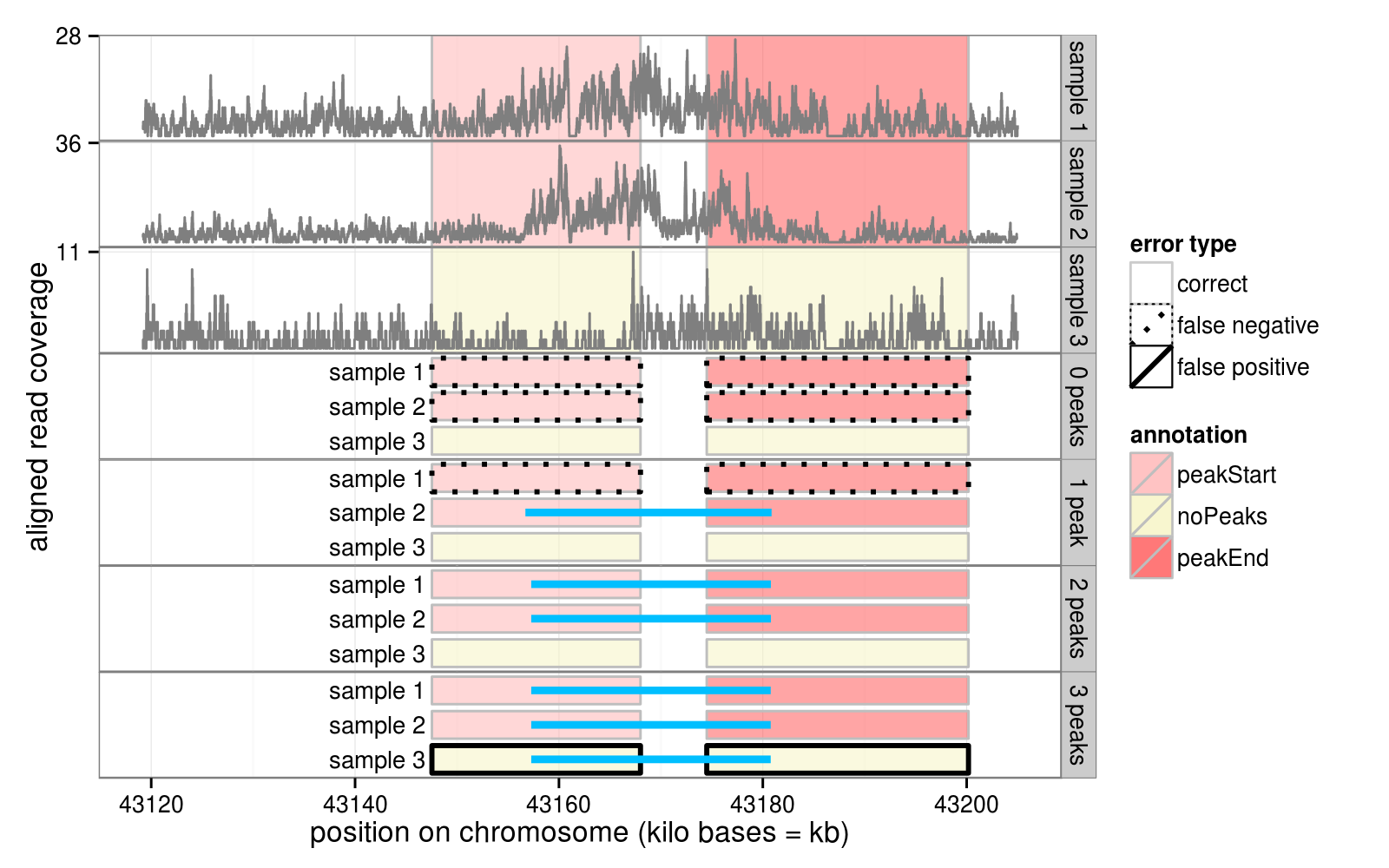}
  \vskip -0.5cm
  \caption{A labeled data set with $S=3$ samples (top 3 panels) and
    the \ref{PeakSegJoint} models for $p\in\{0, 1, 2, 3\}$ peaks
    (bottom 4 panels).
}
  \label{fig:PeakSegJoint}
\end{figure}

For $S$ sample profiles $\mathbf z_1, \dots, \mathbf z_S\in\ZZ_+^B$
defined on the same $B$ bases, we stack the vectors into a matrix
$\mathbf Z\in\ZZ_+^{B \times S}$ of count data. Consider the following
model for the mean matrix $\mathbf M\in\RR^{B\times S}$ which allows
each sample to have either 0 or 1 peaks, with a total of $p\in\{0,
\dots, S\}$ peaks:
\begin{align}
  \label{PeakSegJoint}
  \mathbf{\hat M}^p(\mathbf Z)  =
  \argmin_{\mathbf M\in\RR^{B\times S}} &\ \ 
  \sum_{s=1}^S 
  \text{PoissonLoss}(\mathbf m_s, \mathbf z_s) 
  \tag{\textbf{PeakSegJoint}}
  \\
  \text{such that} &\ 
  \forall s\in\{1, \dots, S\},\, 
  \Peaks(\mathbf m_s)\in\{0, 1\},  
  \label{zero_or_one}
  \\
  &\ 
  \forall s\in\{1, \dots, S\},\,
  \forall j\in\{1, \dots, B\},\, P_j(\mathbf m_s) \in\{0, 1\},
  \label{joint_up_down}
  \\
  &\ 
  p = \sum_{s=1}^S \Peaks(\mathbf m_s)
  \label{total_peaks}
  \\
  &\ \forall s_1\neq s_2\mid
  \nonumber
  \Peaks(\mathbf m_{s_1})=\Peaks(\mathbf  m_{s_2})=1,\,
  \forall j\in\{1, \dots, B\},\\
  &\ \ P_j(\mathbf m_{s_1}) = P_j(\mathbf m_{s_2}).
  \label{joint_constraint}
\end{align}
The first two constraints are similar to the \ref{PeakSeg} model
constraints. The constraint on the number of peaks per sample
(\ref{zero_or_one}) means that each sample may have either zero or one
peak. The constraint (\ref{joint_up_down}) requires the segment mean
$\mathbf m_s$ of each sample to have alternating changes (up, down,
up, down and not up, up, down). The overall constraint
(\ref{total_peaks}) means that there is a total of $p$ samples with
exactly 1 peak. The last constraint (\ref{joint_constraint}) means
that peaks should occur in the exact same positions in each sample.
The \ref{PeakSegJoint} model for a data set with $S=3$ samples is
shown in Figure~\ref{fig:PeakSegJoint}.

\subsection{Supervised penalty learning}

In the last section we considered data $\mathbf Z\in\ZZ_+^{B\times S}$
for $S$ samples in a single genomic region with $B$ bases. Now assume
that we have data $\mathbf Z_1,\dots, \mathbf Z_n\in\ZZ_+^{B\times S}$
for $n$ genomic regions, along with annotated region labels
$L_1,\dots, L_n$. 

For each genomic region $i\in\{1,\dots,n\}$ we can compute a sequence
of \ref{PeakSegJoint} models $\mathbf{\hat M}^0(\mathbf Z_i),\dots,
\mathbf{\hat M}^S(\mathbf Z_i)$, but how will we predict which of
these $S+1$ models will be best?
This is the segmentation model selection problem, which we propose to
solve via supervised learning of a penalty function.

First, for a positive penalty constant $\lambda\in\RR_+$, we define
the optimal number of peaks as
\begin{equation}
  \label{eq:optimal_segments}
  p^*(\lambda, \mathbf Z) =
  \argmin_{p\in\{0, \dots, S\}}
  p \lambda + 
  \text{PoissonLoss}\left[
    \mathbf{\hat M}^p(\mathbf Z),
    \mathbf Z
  \right].
\end{equation}
Also suppose that we can compute $d$-dimensional sample-specific
feature vectors $\mathbf x\in\RR^d$ and stack them to obtain feature
matrices $\mathbf X_1,\dots, \mathbf X_n\in\RR^{d\times S}$. We will
learn a function $f:\RR^{d\times S}\rightarrow\RR$ that predicts
region-specific penalty values $f(\mathbf X_i) = \log \lambda_i\in\RR$
for each genomic region $i$. In particular we will learn a weight
vector $\mathbf w\in\RR^d$ in a linear function $f_{\mathbf w}(\mathbf X) =
\mathbf w^\intercal \mathbf X \mathbf 1_S$, where $\mathbf 1_S$ is a
vector of $S$ ones.

For supervision we use the annotated region labels $L_i$ to compute
the number of incorrect regions (\ref{eq:error}) for each model size
$p$, and then a target interval $\mathbf y_i = ( \underline y_i,
\overline y_i )$ of penalty values \citep{HOCKING-penalties}.
Briefly, a predicted penalty in the target interval $f(\mathbf
X_i)\in\mathbf y_i$ implies that the \ref{PeakSegJoint} model with
$p^*\left[\exp f(\mathbf X_i), \mathbf Z_i\right]$ peaks achieves the
minimum number of incorrect labels $L_i$, among all $S+1$
\ref{PeakSegJoint} models for genomic region $i$.

A target interval $\mathbf y$ is used with the squared hinge loss
$\phi(x)=(x-1)^2 I(x\leq 1)$ to define the surrogate loss
\begin{equation}
  \label{eq:surrogate_loss}
  \ell\left[
    \mathbf y,\,
    \log \hat \lambda
    \right]
    =
    \phi\big[
      \log\hat\lambda - \underline y
    \big]
    +
    \phi\big[
    \overline y - \log\hat\lambda
    \big],
\end{equation}
for a predicted penalty value $\hat \lambda\in\RR_+$. For a weight parameter
$\mathbf w\in\RR^d$, the convex average surrogate loss is
\begin{equation}
  \label{eq:average_surrogate}
  \mathcal L(\mathbf w) =
  \frac 1 n
  \sum_{i=1}^n
  \ell\left[
    \mathbf y_i,\,
     f_{\mathbf w}( \mathbf X_i )
    \right].
\end{equation}
Finally, we add a 1-norm penalty to regularize and encourage a sparse
weight vector, thus obtaining the following convex supervised penalty
learning problem:
\begin{equation}
  \label{argmin_w}
  \mathbf{\hat w}^\gamma = 
  \argmin_{\mathbf w\in\RR^d}
  \mathcal L(\mathbf w) + \gamma ||\mathbf w||_1.
\end{equation}

To predict on test data $\mathbf Z$ with features $\mathbf X$, we
compute the predicted penalty $\hat \lambda = \exp f_{\mathbf{\hat
    w}}(\mathbf X)$, the predicted number of peaks $\hat p = p^*(\hat
\lambda, \mathbf Z)$, and finally the predicted mean matrix
$\mathbf{\hat M}^{\hat p}(\mathbf Z)$. Each column/sample of the mean
matrix has either two changes (the second segment is the peak) or no
changes (no peak).

\section{Algorithms}
\label{sec:algorithms}

\subsection{Heuristic discrete optimization for joint segmentation}

The \ref{PeakSegJoint} model is defined as the solution to an
optimization problem with a convex objective function and non-convex
constraints.  Real data sets $\mathbf Z\in\ZZ_+^{B\times S}$ may have
a very large number of data points to segment $B$. Explicitly
computing the maximum likelihood and feasibility for all $O(B^2)$
possible peak start/endpoints is guaranteed to find the global
optimum, but would take too much time.  Instead, we propose to find an
approximate solution using a new discrete optimization algorithm
called \JointHeuristic~(Algorithm~1).

\newpage

The main idea of the \JointHeuristic\ algorithm is to first zoom out
(downsample the data) repeatedly by a factor of $\beta$, obtaining a
new data matrix of size $b\times S$, where $b \ll B$
(line~\ref{zoomout}). Then we solve the \ref{PeakSegJoint} problem for
$p$ peaks via \textsc{GridSearch} (line~\ref{gridsearch}), a
sub-routine that checks all $O(b^2)$ possible peak start and end
positions. Then we zoom in by a factor of $\beta$ (line~\ref{zoomin})
and refine the peak positions in $O(\beta^2)$ time via
\textsc{SearchNearPeak} (line~\ref{searchnear}). After having zoomed
back in to the BinSize=1 level we return the final Peak positions, and
the $p$ Samples in which that peak was observed.

\begin{figure}[b!]
  \centering
  \includegraphics[width=0.95\textwidth]{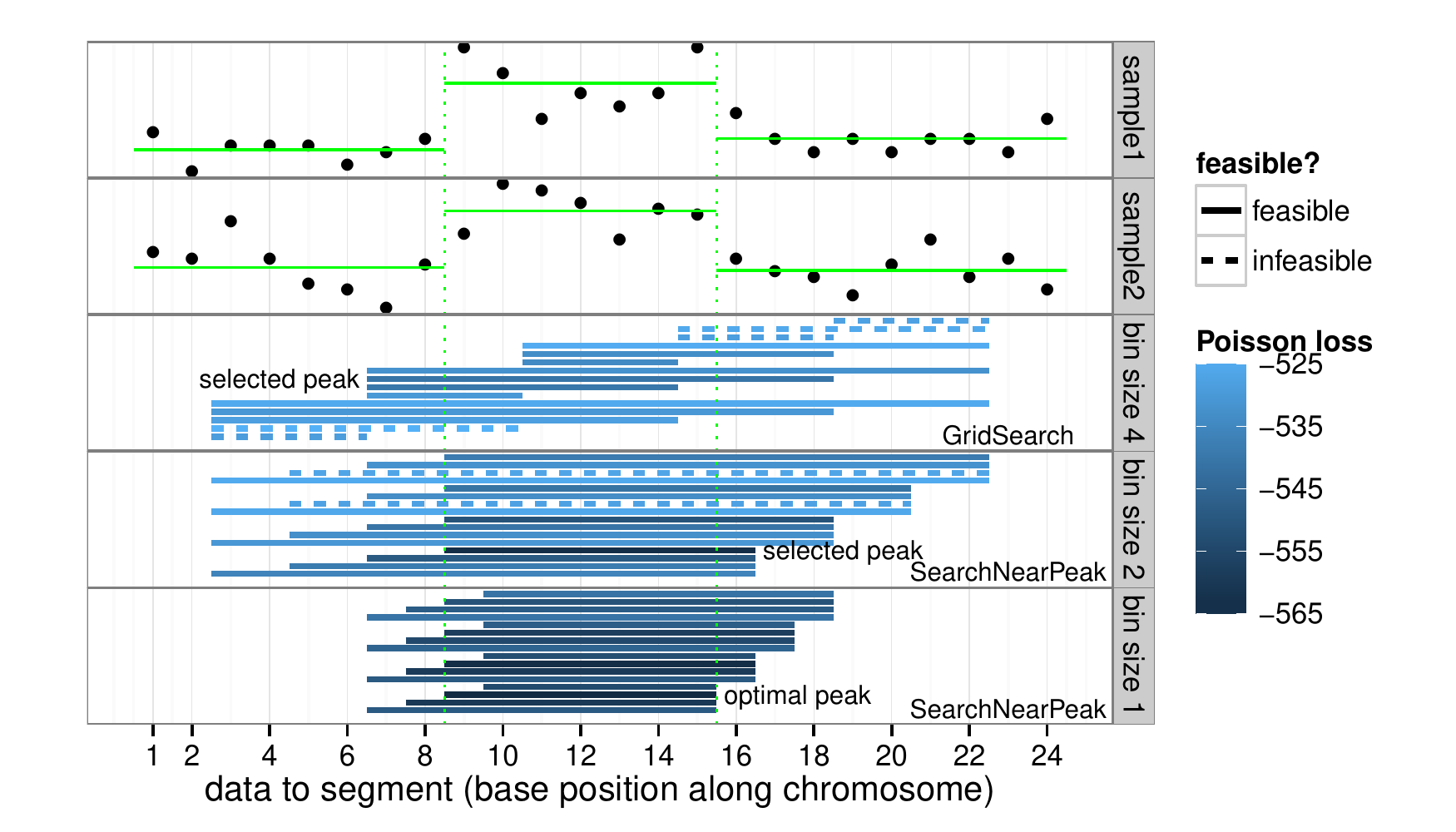}
  \vskip -0.5cm
  \caption{Demonstration of the \JointHeuristic\ segmentation
    algorithm. For a data set with $B=24$ data points and $S=2$
    samples (top 2 panels), the algorithm with zoom factor of
    $\beta=2$ proceeds as follows. First, the Poisson loss and
    feasibility (\ref{joint_up_down}) is computed via
    \textsc{GridSearch} over all peak starts and ends at bin size
    4. The feasible model with minimum Poisson loss is selected, the
    bin size is decreased to 2, and \textsc{SearchNearPeak} considers
    a new set of models around the selected peak starts and ends. We
    continue and return the optimal model at bin size 1 (shown in
    green). Interactive figure at \url{http://bit.ly/1AA6TgK}}
  \label{fig:heuristic-algo}
\end{figure}

\begin{algorithm}[H]
\begin{algorithmic}[1]
  \REQUIRE count data $\mathbf Z\in\ZZ_+^{B\times S}$, number of
  peaks $p\in\{0, \dots, S\}$, zoom factor
  $\beta\in\{2, 3, \dots\}$.
  \STATE $\textrm{BinSize} \gets \textsc{MaxBinSize}(B, \beta)$. \label{zoomout}
  \STATE $\textrm{Peak}, \textrm{Samples} \gets \label{gridsearch}
  \textsc{GridSearch}(\mathbf Z, p, \textrm{BinSize})$.
  \WHILE{$1 < \textrm{BinSize}$}
  \STATE $\textrm{BinSize} \gets \textrm{BinSize} / \beta$. \label{zoomin}
  \STATE $\textrm{Peak} \gets
  \textsc{SearchNearPeak}(\mathbf Z, \textrm{Samples}, \label{searchnear}
  \textrm{BinSize}, \textrm{Peak})$
  \ENDWHILE
  \RETURN Peak, Samples.
  \caption{\JointHeuristic, available at
    \url{https://github.com/tdhock/PeakSegJoint}}
\end{algorithmic}\label{algo}
\end{algorithm}

For example if we fix the zoom factor at
$\beta=2$, a demonstration of the algorithm on a small data set 
with $B=24$ points is shown in
Figure~\ref{fig:heuristic-algo}. \textsc{MaxBinSize} returns 4, so
\textsc{GridSearch} considers 15 models of $b=7$ data points at bin
size 4, and then \textsc{SearchNearPeak} considers 16 models each at
bin sizes 2 and 1. In the real data set of
Figure~\ref{fig:PeakSegJoint}, there are $B=85846$ data points,
\textsc{MaxBinSize} returns 16384, \textsc{GridSearch} considers 10
models of $b=6$ data points, and then \textsc{SearchNearPeak}
considers 16 models each at bin sizes 8192, 4096, ..., 4, 2, 1.

Overall \JointHeuristic\ with zoom factor $\beta$ searches a total of
$O(\beta^2\log B)$ models, and computing the likelihood and
feasibility for each model is an $O(pB)$ operation, so the algorithm
has a time complexity of $O(\beta^2 pB\log B)$. Thus for a count data
matrix $\mathbf Z\in\ZZ_+^{B\times S}$, computing the sequence of
\ref{PeakSegJoint} models $\mathbf{\hat M}^0(\mathbf Z), \dots,
\mathbf{\hat M}^S(\mathbf Z)$ takes $O(\beta^2 S B\log B)$ time.

Finally note that when there is only $S=1$ sample, \JointHeuristic\
can be used to find an approximate solution to the \ref{PeakSeg} model
with $p=1$ peak.

\subsection{Convex optimization for supervised penalty learning}

The convex supervised learning problem (\ref{argmin_w}) can be
solved using gradient-based methods such as FISTA, a Fast Iterative
Shinkage-Thresholding Algorithm \citep{fista}. 
For FISTA with constant step size we also need a Lipschitz constant of
$\mathcal L(\mathbf w)$. Following the arguments of
\citet{hingeSquareFISTA}, we derived a Lipschitz constant of
$\sum_{i=1}^n ||\mathbf X_i||_F^2/n$. Finally, we used the
subdifferential stopping criterion of \citet{HOCKING-penalties}.


\section{Results}
\label{sec:results}

\subsection{Accuracy of PeakSegJoint on benchmark data sets}

We used McGill benchmark data sets from
\url{http://cbio.ensmp.fr/~thocking/chip-seq-chunk-db/} which included
a total of 12,826 manually annotated region labels
\citep{hocking2014visual}. Each data set contains labels
grouped into windows of nearby regions (from 4 to 30 windows per data
set). For each data set, we performed 6 random splits of windows into
half train, half test. Since there are multiple peaks per window, and
\ref{PeakSegJoint} can detect at most 1 peak per sample, we divided
each window into separate \ref{PeakSegJoint} problems of size $B$ (a
hyper-parameter tuned via grid search).

\begin{figure}[H]
  \centering
  \includegraphics[width=\textwidth]{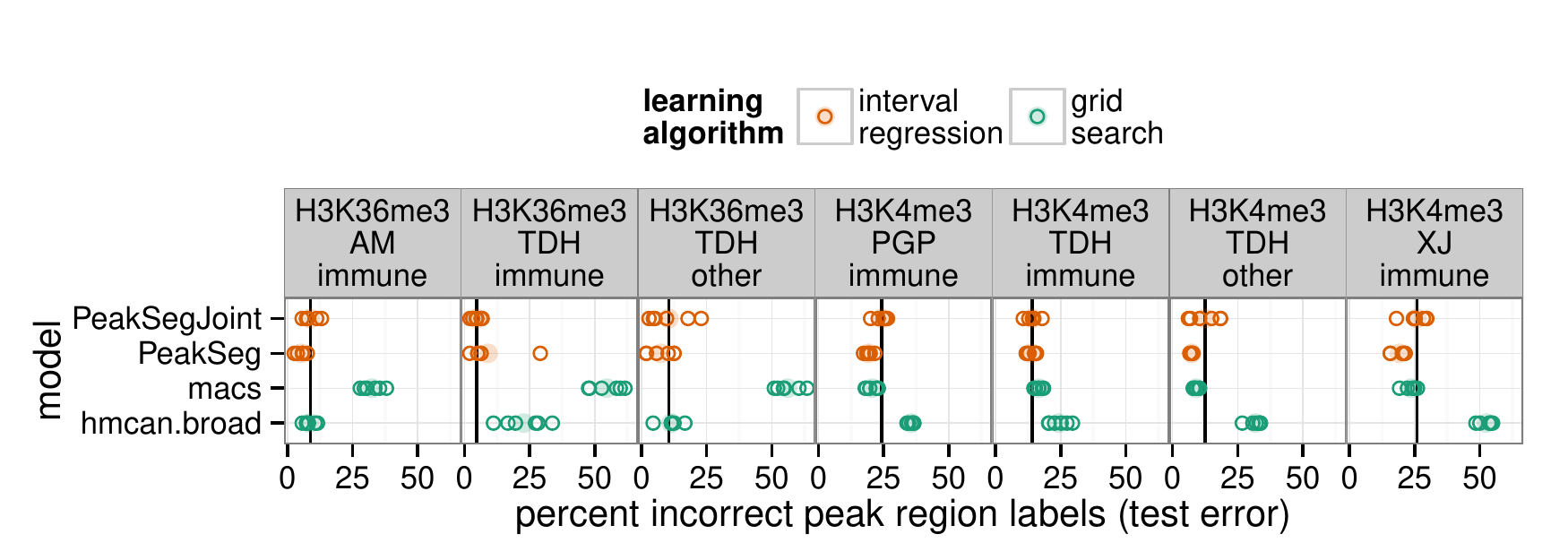}
  \vskip -0.5cm
  \caption{Test error of peak detectors in the 7 McGill 
    benchmark data sets (panels from left to right). Each dot shows
    one of 6 train/test splits, and the black vertical line marks the
    mean of the \ref{PeakSegJoint} model.}
  \label{fig:test-error-dots}
\end{figure}

We compared the test error of \ref{PeakSegJoint} with three
single-sample models (Figure~\ref{fig:test-error-dots}). In all 7 of
the McGill benchmark data sets, our proposed \ref{PeakSegJoint} model
achieved test error rates comparable to the previous state-of-the-art
\ref{PeakSeg} model. In contrast, the baseline hmcan.broad
\citep{HMCan} and macs \citep{MACS} models from the bioinformatics
literature were each only effective for a single experiment type
(either H3K36me3 or H3K4me3, but not both).


\newpage

\subsection{Speed of heuristic PeakSegJoint algorithm}

\begin{figure}[H]
  \centering
  \input{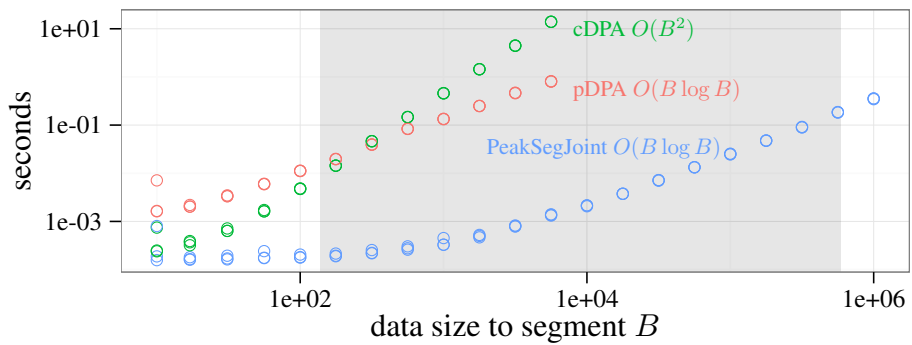}
  \vskip -0.5cm
  \caption{Timings of three Poisson segmentation algorithms on
    simulated data sets of varying size $B$. The grey shaded area
    represents the range of problem sizes selected in the McGill
     benchmark data sets. }
  \label{fig:timings}
\end{figure}

The \JointHeuristic\ algorithm is orders of magnitude faster than
existing Poisson segmentation algorithms
(Figure~\ref{fig:timings}). For simulated single-sample, single-peak
data sets of size $B\in\{10^1, \dots, 10^6\}$, we compared
\JointHeuristic\ with the $O(B\log B)$ unconstrained pruned dynamic
programming algorithm (pDPA) of \citet{Segmentor} (R package
Segmentor3IsBack), and the $O(B^2)$ constrained dynamic programming
algorithm (cDPA) of \citet{HOCKING-PeakSeg} (R package PeakSegDP). We
set the maximum number of segments to 3 in the pDPA and cDPA, meaning
one peak. Figure~\ref{fig:timings} indicates that the
\JointHeuristic\ algorithm of \ref{PeakSegJoint} enjoys the same $O(B
\log B)$ asymptotic behavior as the pDPA. However the \JointHeuristic\
algorithm is faster than both the cDPA and pDPA by at least two orders
of magnitude for the range of data sizes that is typical for real data
($10^2 < B < 10^6$).

These speed differences are magnified when applying these Poisson
segmentation models to real ChIP-seq data sets. For example, the hg19
human genome assembly has about $3\times 10^9$ bases. For the
H3K36me3\_AM\_immune data set, a resolution of about 200 kilobases per
problem was optimal, so running PeakSegJoint on the entire human
genome means solving about 15,000 segmentation problems. PeakSegJoint
takes about 0.1 seconds to solve each of those problems
(Figure~\ref{fig:timings}), meaning a total computation time of about
25 minutes. In contrast, using the pDPA or cDPA would be orders of
magnitude slower (hours or days of computation).

\newpage

\section{Discussion}
\label{sec:discussion}

The table below presents a summary of our proposed \ref{PeakSegJoint}
model alongside several other peak detection algorithms. Time
complexity is for detecting one peak in $B$ data points. The ``?''  of
the unsupervised joint peak detectors means that their articles did
not discuss time complexity \citep{MACS, HMCan, JAMM, PePr}.

\begin{center}
  \begin{tabular}{crrcc}
  Algorithm & Samples & Cell types & Time complexity & Learning \\
  \hline
  macs & 1 & 1 & ? & unsupervised \\
  hmcan.broad & 1 & 1 & ? & unsupervised \\
  PeakSeg (cDPA) & 1 & 1 & $O(B^2)$ & supervised \\
  JAMM & $\geq 1$ & 1 & ? & unsupervised \\
  PePr & $\geq 2$ & 1--2 & ? & unsupervised \\
  PeakSegJoint (JointZoom) & $\geq 1$ & $\geq 1$ & $O(B \log B)$ & supervised
\end{tabular}
\end{center}

\subsection{Multi-sample PeakSegJoint versus single-sample PeakSeg}

The \ref{PeakSegJoint} model is explicitly designed for supervised,
multi-sample peak detection problems such as the McGill benchmark data
sets that we considered. We observed that the single-sample
\ref{PeakSeg} model is as accurate as \ref{PeakSegJoint} in these data
(Figure~\ref{fig:test-error-dots}). However, any single-sample model
is qualitatively inferior to a multi-sample model since it can not
predict overlapping peaks at the exact same positions. Furthermore,
the $O(B \log B)$ \JointHeuristic\ algorithm can compute the
\ref{PeakSegJoint} model much faster than the $O(B^2)$ cDPA of
\ref{PeakSeg} (Figure~\ref{fig:timings}).

\subsection{PeakSegJoint versus other joint peak detectors}

We attempted to compare with the multi-sample JAMM and PePr
algorithms. The \ref{PeakSegJoint} model is more
interpretable than these existing methods, since it is able to handle
several sample types. This was advantageous in data sets such as
Figure~\ref{fig:good-bad}, which contains 3 cell types: tcell, bcell,
and monocyte. The PeakSegJoint model can be run once on all cell
types, and the resulting model can be easily interpreted to find the
differences, since peaks occur in the exact same locations across
samples. In contrast, existing methods such as JAMM or PePr are less
suitable to analyze this data set since they are limited to modeling
only 1 or 2 cell types. We nevertheless downloaded and ran the
algorithms separately on each sample type, but both JAMM and PePr
produced errors and so were unable to analyze the McGill benchmark
data sets.

\section{Conclusions}
\label{sec:conclusions}

We proposed the \ref{PeakSegJoint} model for supervised joint peak
detection. It generalizes the state-of-the-art \ref{PeakSeg} model to
multiple samples. We proposed the \JointHeuristic\ Poisson
segmentation algorithm that is orders of magnitude faster than
existing algorithms. Finally, we showed that choosing the number of
peaks in \ref{PeakSegJoint} using supervised penalty learning yields
test error rates that are comparable to the previous state-of-the-art
\ref{PeakSeg} model.

For future work, we would be interested in a theoretical analysis
analogous to the work of \citet{cleynen2013segmentation}, which could
suggest the form of the optimal penalty function to choose the number
of peaks in the \ref{PeakSegJoint} model.

\newpage

\bibliographystyle{abbrvnat}

\bibliography{refs}

\end{document}